\documentclass{article}

\usepackage{arxiv}

\usepackage[utf8]{inputenc} 
\usepackage[T1]{fontenc}    
\usepackage{hyperref}       
\usepackage{url}            
\usepackage{booktabs}       
\usepackage{amsfonts}       
\usepackage{nicefrac}       
\usepackage{microtype}      
\usepackage{lipsum}		
\usepackage{graphicx}
\usepackage[square,numbers]{natbib}
\usepackage{doi}
\usepackage{amsmath}
\usepackage{amssymb}

\title{SDA-GAN:Unsupervised Image Translation Using Spectral Domain Attention-Guided Generative Adversarial Network}

\author{ \href{https://orcid.org/0000-0003-4883-1068}{\includegraphics[scale=0.06]{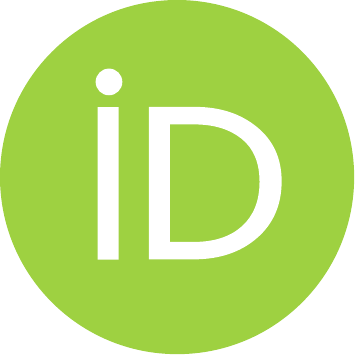}\hspace{1mm}Qizhou Wang}) \\
	Department of CEMSE\\
	King Abdullah University of Science and Technology\\
	Thuwal 23955 \\
	\texttt{qizhou.wang@kaust.edu.sa} \\
	\And
	\href{https://orcid.org/0000-0000-0000-0000}{\includegraphics[scale=0.06]{orcid.pdf}\hspace{1mm}Maksim Makarenko} \\
    Department of CEMSE\\
	King Abdullah University of Science and Technology\\
	Thuwal 23955 \\
	\texttt{maksim.makarenko@kaust.edu.sa} \\
}

\renewcommand{\shorttitle}{\textit{arXiv} Template}

\hypersetup{
pdftitle={SDA-GAN},
pdfsubject={CV},
pdfauthor={Qizhou Wang},
pdfkeywords={Image translation, Generative network, Computer vision},
}

\begin{document}
\maketitle

\begin{abstract}
	This work introduced a novel GAN architecture for unsupervised image translation on the task of face style transform. A spectral attention based mechanism is embedded into the design along with spatial attention on the image contents. We proved that neural network has the potential of learning complex transformations such as Fourier transform, within considerable computational cost. The model is trained and tested in comparison of the baseline model, that only uses spatial attention. The performance improvement of our approach is significant especially when the source and target domain include different complexity (reduced FID to 49.18 from 142.84). In the translation process, a spectra filling effect was introduced due to the implementation of FFT and spectral attentions. Other style transfer tasks and real-world object translation is also studied in this paper.
\end{abstract}


\section{Introduction}

Human has the nature of self-expressing, however the fear of total exposure to the outer world still remains deep inside our mind. In this situation, some simple alienation on their outlook may satisfy such ambivalence. People are using all kinds of decoration and filters on their selfies, before they post them on the internet. As the extensive use of social media in entertainment industry keeps arising, style filters have become the base function of many applications.

Recently, Generative Adversarial Networks (GANs) \cite{goodfellow2014generative} have shown the potential of image-to-image translation automatically. Differing from the original GAN, the state-of-art approaches (eg: CycleGAN\cite{zhu2017unpaired}, Pix2PixHD\cite{wang2018high}, StyleGAN\cite{karras2019style}, DRIT++\cite{lee2020drit++}, ) implement the discriminator to distinguish both the generated image and the image from desired domain. Meanwhile, the generator is forced to output images with similar features in this domain. As great results were produced by these methods. In the unsupervised task, the high-level semantic information in a picture was still poorly engineered. The intuitive solution to this problem is to have pre-trained semantic segmentation masks and implement them into the generator layers. ContrastGAN \cite{liang2017generative} uses manual object-mask annotations to disentangle the foreground and background of the existing picture. However, even a well-prepared segmentation model would cause loss of generalization of the network \cite{alotaibi2020deep} due to the hand-labeled data has little scale and availability. Thus, recent years have seen many types of research based on the self-attention of GANs \cite{chen2018attention, tang2019attentiongan,zhang2019self, emami2020spa}. This technique allows the generator and discriminator to extract the latent attention mask of the input image, according to the learned data distribution. In image-to-image translation practices, the attention mechanism is embedded into the GAN architecture by branching convolutional blocks of the image encoder/decoder. The spatial attention map (implemented by dot product) is used to distinguish foreground object and background noise, thus excludes the background from translation. These methods largely improve the image quality, when dealing with translation between real-world objects like human faces. However spatial attention is not sufficient to cover the low-level features such as artwork movement, texture, due to the style change in 2 groups of images that are global mapping. In this work, we focus on the translation between Marvel-styled comic image and human face, which include both high-level semantic manipulation and low-level feature engineering. 

In order to address the existing issues, we propose a novel Spectral Domain Attention-Guided Generative Adversarial Network (SDA-GAN) for unsupervised image-to-image translation. As shown in Fig.\ref{sda}, by adding spectral attention with spatial attention, the model would better translate the low-level features, while still having objective consistency. A two-scheme framework is applied in this work. First, we explore the most straightforward method to inject spectral information into the network architecture. We operate Fast Fourier Transform directly on the content layer of the generator. Then we combine the contents with the spectral attention and spatial attention with inverse Fourier transform. Inspired by \cite{tang2019attentiongan}, the attention is divided into multiple foreground and background parts for each case. The generator produces a transform map, which includes both objective content and global style. As a proof of concept, we suppose the network will eventually learn how to distinguish the object and background, after fine-tuning and optimization. Thus the final output can be divided into two components: foreground (generated by foreground attention and content/style mask) and background (generated by background attention and original image). By weighted-adding the two counterparts, we obtain the mapping from the source domain to the target domain. This transformation includes both spatial modification and global texture, styles, illumination, etc. Due to introducing semantic operations, the uninterested area of images can be ignored, during the translation. To reduce the computational cost introduced by complex number operation, we divided the spectral information into two parts: phase and amplitude. The phase is shared in all channels, while all the generator layers only have to produce amplitude content and attention. This pre-separation of phase and amplitude can help stablizing the network, preventing it from numerical error.

Furthermore, we constructed a more complicated system, where the spectral contents are independently generated from its own branch. Experimentally, we discovered that cycle architecture \cite{zhu2017unpaired} is vital in the training process, for it largely reduces the possibility of mode collapse and preserves global consistency. 

The major contribution of this work is summarized as follows:

\begin{itemize}
  \item To the extent of our knowledge, this is the first work that implemented spectral attention on image translation.
  \item We propose a novel spectral attention-guiged generator that is robust both on high-level features and low-level features. This architecture utilizes spectral and spatial information, which allows the network to perform local translating and global style reformatting on the objectives while ignoring backgrounds.
  \item We demonstrate that the use of Fourier transform in convolutional networks will introduce performance benefits, while only slightly impairs the computational efficiency.
  \item We introduce an adaptive image fuser, that integrates translation contents from spatial and spectral domain.
\end{itemize}

\section{Related Work}
\textbf{Generative Adversarial Networks} coined in 2014 by Goodfellow et al.\cite{goodfellow2014generative} have achieved significant results on various computer vision subfields, including super-resolution \cite{dahl2017pixel,bulat2018learn}, image-to-image translation \cite{wang2018high,karras2019style,lee2020drit++}, text-to-image generation \cite{reed2016generative}, and semantic segmentation \cite{cherian2019sem,yao2018exploring}. 
A GAN is composed of two confrontational networks called Generator G and Discriminator D, which are trained simultaneously using the adversarial loss. G tries to generate sufficiently high-quality fake images and D is trained to distinguish them from real pictures. The ultimate goal of training a GAN is to achieve Nash equilibrium, as the enhancement of one part in the network would cause loss of the counterpart. In this case, we hope that the generator produces images with high confidence both in discriminators and human eyes. Original GANs derive images from random vectors, while conditional GANs (cGANs) utilize external information such as annotations, interested points, semantic mask and desired image domain. In this work, we adopt the adversarial loss to learn the mapping from the source domain to the target domain so that the translated image will be genuine enough as the ground truth, from the perspective of discriminators.

\textbf{Image-to-Image translation} 
Using a convolutional neural network (CNN), GANs can learn and perform a translation of images from two distinct domains (eg: horses to zebras, summers to winters, cry to laughter). Recent years have seen a variety of applications with such architecture \cite{isola2017image,wang2018high,reed2016generative}. For paired dataset, each image in domain A has a corresponding image in domain B. The translation between these domains has been widely explored. Taking a classic example, Pix2Pix \cite{isola2017image} learns a mapping using conditional network with L1 loss to stabilize the convergence and reduce blurring. The further Pix2PixHD \cite{wang2018high} increased the resolution limit to 2048*1024 by implementing a pyramid of generators, which produce images from coarse to fine quality. Pix2PixHD enables dynamic editing of image semantics with labels or segmentation masks in a short processing time. 

However, the issue of limited training samples hinders the application of paired image-to-image translation. To overcome the limitation, several methods were proposed to handle unpaired image-to-image translation. CycleGAN \cite{zhu2017unpaired} is the ancestor of all cycle-consistency-based GANs. They preserve the key features in the whole image processing stage, by adding cycle-consistency loss and identity loss to the adversarial loss. The variants of CycleGAN include Augmented CycleGAN \cite{almahairi2018augmented}, DiscoGAN \cite{kim2017learning}, DualGAN \cite{yi2017dualgan}, Attention GAN \cite{tang2019attentiongan} and etc. CycleGAN builds a symmetric architecture involving two generators and discriminators and trained them recurrently, giving the training image unchanged. We also inherit this cycle-based architecture in our network, and the detailed illustration is in Section 3. Besides cycle-based networks, autoencoder-based GANs \cite{liu2016coupled, liu2020gmm} and Disentangled Representation methods \cite{huang2018multimodal,lee2018diverse,lee2020drit++} are also widely investigated.

\textbf{Attention-Guided GANs}
Imitating the human vision system, attention mechanism has been widely applied in many fields of computer vision including visual explanation \cite{fukui2019attention}, semantic segmentation \cite{chen2016attention}, image and video captioning \cite{yan2019stat}, etc. Attention stimulates the network to focus on the region of interest, thus improve the performance of all the above applications. 

Recently the incorporation of attention and GANs has been extensively studied. These approaches can be divided into 2 classes. The first category is GAN-assisted attention generation. In \cite{chen2016attention,yao2018exploring,han2018spine}, GANs are employed to complete the segmentation task. As given paired images and semantic masks, the GANs are trained to build the mapping between these two domains. The second category is attention-guided GANs. In ContrastGAN proposed in 2018, the authors used a mask-conditional generative model to disentangle the object and background. The network was trained on datasets with extra segmentation masks which largely limited the application fields. To address this issue, SAGAN \cite{zhang2019self} is proposed to combine attention mechanism and GAN to deal with unlabeled data. The self-attention network is pre-trained in a segmentation dataset and then serves as a teacher network that inserts attention masks into the generator and discriminator. The system can be trained in an end-to-end manner and can process new data without extra information.

However, even a well-prepared segmentation model would cause loss of generalization of the network \cite{alotaibi2020deep} due to the hand-labeled data is always hard to attain. Thus, many types of research based on the integrated self-attention of GANs \cite{chen2018attention, tang2019attentiongan, emami2020spa} have been studied. This technique allows the generator to extract the latent attention mask of the input image, according to the learned data distribution. Instead of using the labeled segmentation mask, the network gradually learns the attention during the generation-discrimination iteration. 

By fusing the attention mask and generated content, the generated image will only contain translations in the desired features, leaving the background or shared features unchanged. This avoids artifacts of the image, compared to the original GAN with no attention information. Popular attention models take advantage of the spatial distribution of interested regions. In this work, we expand the attention to the spectral domain. Efficiently applying Fourier transform and inverse Fourier transform in the network architecture, we demonstrate that spectral-domain attention can improve the generator performance. The enhanced awareness of desired translation could be further applied in other image representation fields.

\section{SDA-GAN}
We first formulate the image-to-image translation problem as learning the mapping $G_X$ from a source image $x\in X$ to the target image $y\in Y$. The discriminator $D_Y$ reacts to the generated images and real sample to predict a confidence score of each type. To preserve cycle consistency, a backward translation network with generator $G_Y$ and discriminator $D_x$ is built for translation from $Y$ to $X$. The adversarial loss $L_{adv}$ indicates the accuracy of the classification. Based on the loss function, the two parts of GAN are trained to play a mini-max game. For generator, the goal is to fool the discriminator with fake images so the object will be maximizing $L_{adv}$. For discriminator, the goal is to distinguish precisely whether an image is real or generated so the object will be minimizing $L_{adv}$. At the very beginning of training, $G_X$ can only produce noise-like images, that are easily distinguished by $D_Y$. 
After sufficient iterations, the optimal situation is that $G_X$ produces nice images that look like real ones, and $D_Y$  has 50\% accuracy of identifying real and fake images. In the perspective of optimization, a saddle point is achieved, which implies the instability of GANs, that no global optimization point exists. The instability is a major issue in training GANs, while the previous study has proved mathematically that by modifying and constraining the loss function and weight distribution, local equilibrium is achievable for GANs \cite{ kodali2017convergence}. In this work, two configurations of spectral engineering are constructed and tested.

\subsection{Spectral Attention Guided Generator}
\begin{figure*}
    \centering
    \includegraphics[width=0.9\textwidth]{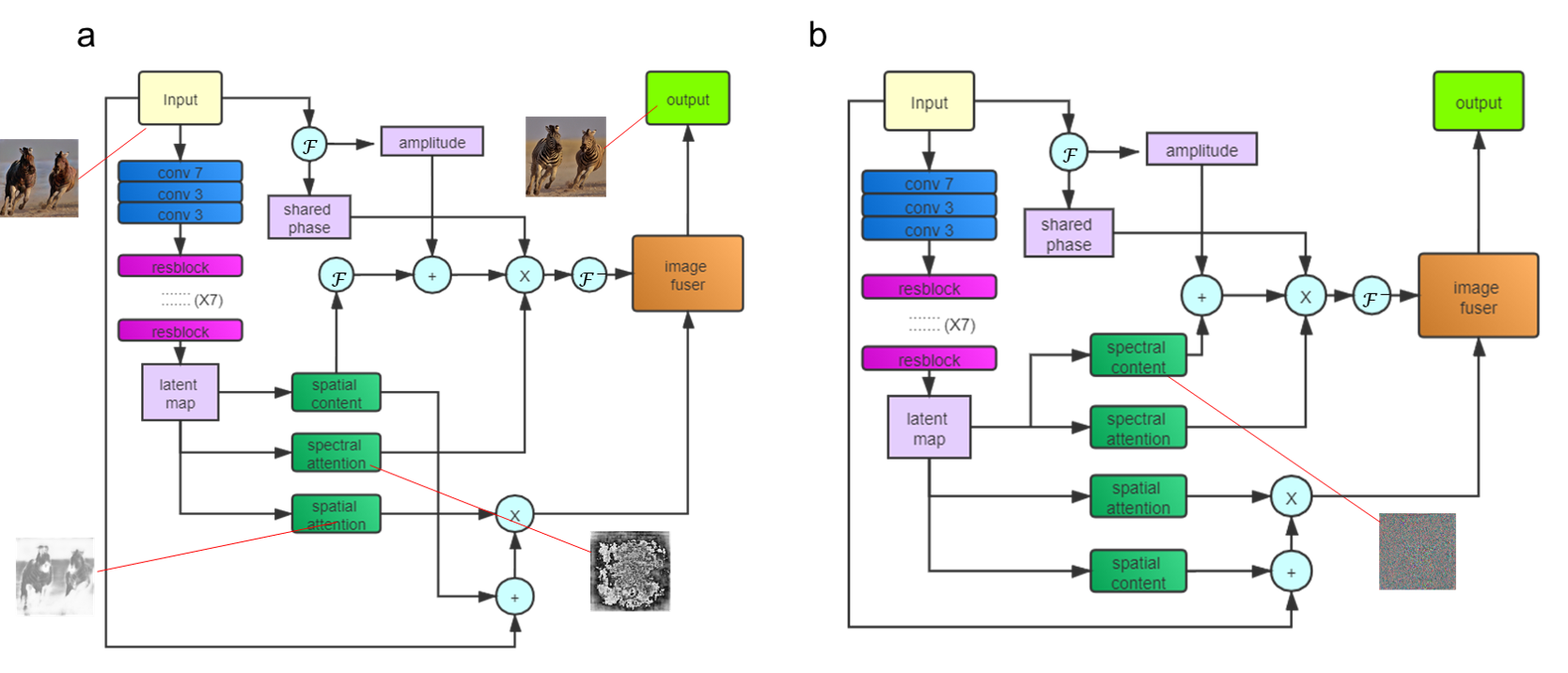}
    \caption{\textbf{(a)} Diagram of basic architecture in SDA-GAN. \textbf{(b)} Diagram of an alternative architecture in SDA-GAN}
    \label{sda}
\end{figure*}

In the proposed network, we use $A_X$ and $A_Y$ to denote the spatial attention, as well as $S_X$ and $S_Y$ for spectral attention of Y-X and X-Y mapping. $C_X$ and $C_Y$ are generated contents, sorted channel-wise. Then, the objective mapping can be represented as $G(x)=G(A_y, S_y, C_y) ,H(y)=H(A_x,S_x,C_x)$ . As shown in Fig.\ref{sda}, the image input x is first encoded into a latent code $E_x$ by convolutional blocks. Then a three-branch network ($G_A$,$G_Y$,$G_S$)    respectively generate the spatial attention $A_y$, translation contents $C_y$, and spectral attention $S_y$ using the latent code $E_x$. Further, the 2-D Fourier transform of generated contents $\mathcal{F}(C_x)$ and $\mathcal{F}(C_y)$ are calculated and stored.

In the first configuration described in Fig.\ref{sda}(a), since the shared encoder compress the shape of $x$, both contents and attention decoder can be constructed and trained in an efficient way. To achieve good performance on complex objects, we applied multiple layers of attention extractors $A_y \in \mathbb{R} ^ {H \times W \times n}$ and  $S_y \in \mathbb{R} ^ {H \times W \times n}$, where $H$ and $W$ is the height and width of the feature map, to fine-split the interested region into a sequence of attention masks. Softmax activation is applied as the top layers of $G_A$ and $G_S$, this modification ensures the attention among each channel is a probabilistic distribution. The corresponding translation content has the dimension $C_y \in \mathbb{R} ^ {H \times W \times n \times 3}$, for the generated image should contain RGB channels.

As for spectral features, the translation contents (amplitude) are directly extracted from the output of spatial content decoder, which can be represented as the amplitude of 2D-FFT: $abs(\mathcal{F}(C_x))$. For each channel, the amplitude is combined with shared phase form input image. Finally, a dot-product operation is performed on all paired attention and contents.

To better manipulate the distribution of generated image, an additional convolutional block is employed, as an 'image fuser'. The two branches of generated contents will be the input of image fuser, where spatial and spectral features are processed and integrated, to produce the final output. This implementation gives the network ability to adaptively fine-tune the color theme of generated images. The forward process is presented below:

\begin{equation}
G(x) = N_y(\sum\limits_{i = 1}^n {\left( {\lambda _{Ay}^iC_y^i*A_y^i + \lambda _{Sy}^i({\mathcal{F}^{ - 1}}(S_y^i*\mathcal{F}(C_y^i))} \right)} )
\end{equation}

where $N_y$ denotes for the image fuser. For $i \ne n$, $A_y^i$ and $S_y^i$ are foreground attentions, $A_y^n$ and $S_y^n$ are the background attentions, all satisfying the following relationships:

\begin{equation}
\sum\limits_i^n {A_y^n = 1} ,\sum\limits_i^n {S_y^n = 1} 
\end{equation}

This indicates the freedom of attention at each pixel is $n-1$. $C_y^i$ is the translation contents in each layer, in which $C_y^n=x$ is the identity mapping, multiplied with the background attention. 
Following the same procedure, the output of mapping $ H(y)=H(A_x,S_x,C_x)$ can be derived by:
\begin{equation}
H(y) = {N_x}(\sum\limits_{i = 1}^n {\left( {\lambda _{Ax}^iC_x^i*A_x^i + \lambda _{Sx}^i({\mathcal{F}^{ - 1}}(S_x^i*\mathcal{F}(C_x^i))} \right)} )
\end{equation}

In the second configuration shown in Fig.\ref{sda}(b), an alternative branch of spatial content is implemented. This structure avoids performing FFT on the spatial content layer, by using an independent spectral decoder to generate the amplitude of each channel $\tilde{C}_x$ and $\tilde{C}_y$. At last, the generated amplitudes, together with the background amplitude (from input) are combined with the shared phase. Experimentally, the two mentioned structures don't differ much at performance. However the training is more stable and with less change to achieve 'nan' with the independent amplitude structure. The forward process is presented below:
\begin{equation}
  G(x) = {N_y}(\sum\limits_{i = 1}^n {\left( {\lambda _{Ay}^iC_y^i*A_y^i + \lambda _{Sy}^i({\mathcal{F}^{ - 1}}(S_y^i*{{\tilde C}^i_y})} \right)} )
\end{equation}

\subsection{Spectral Attention Guided Discriminator}
As the generator output remains the same for both structures mentioned above. The discriminator constructs two different loss functions for each configuration. Same as cGANs, the adversarial loss is calculated as below:
\begin{equation}
    \begin{aligned} \mathcal{L}_{adv}\left(G, D_{Y}\right) =-\mathbb{E}_{y \sim p_{\text {data }}(y)}\left[\log D_{Y}(y)\right] \\ -\mathbb{E}_{x \sim p_{\text {data }}(x)}\left[\log \left(1-D_{Y}(G(x))\right)\right] \end{aligned}
\end{equation}
The update direction of weights can be calculated using back-propagation algorithm as generator intend to maximize the loss while discriminator minimizes it. Replacing $G$ with $H$, $D_Y$ with $D_X$, it gives the adversarial loss function of mapping $ H(y)=H(A_x,S_x,C_x)$, for the first configuration, and $ H(y)=H(A_x,S_x,C_x,\tilde{C_x})$ for the second configuration. 
\begin{figure*}[t]
    \centering
    \includegraphics[width=0.9\textwidth]{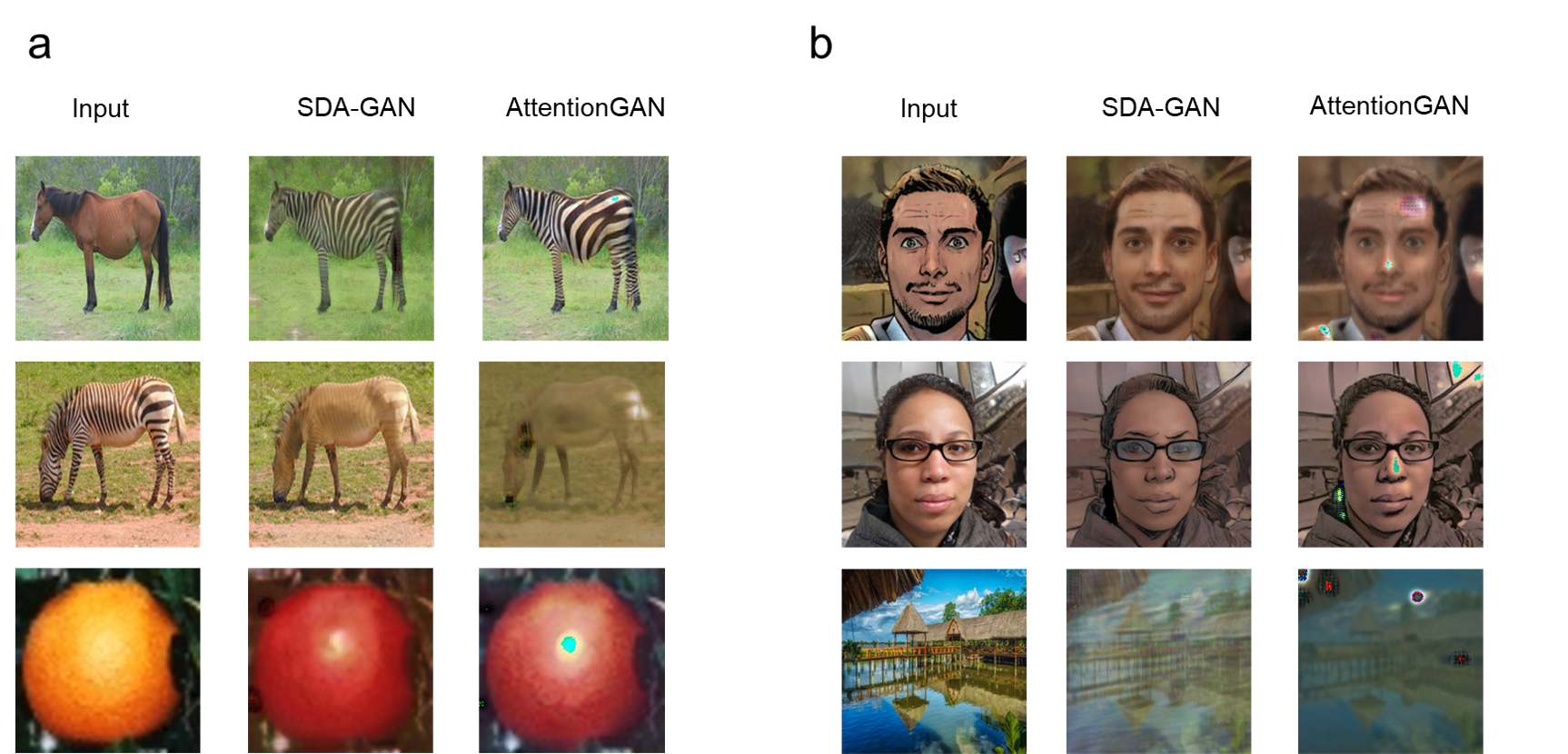}
    \caption{Image translation results of \textbf{(a)} Natural objects and \textbf{(b)} artistic styles}
    \label{res}
\end{figure*}

\begin{table*}
\caption{Performance evaluation of SDA-GAN and baseline}
  \centering
\begin{tabular}{@{}lllllllll@{}}
\toprule
        & \multicolumn{2}{c}{face2comic}                          & \multicolumn{2}{c}{horse2zebra}                         & \multicolumn{2}{c}{apple2orange}                        & \multicolumn{2}{c}{monet2photo}                         \\ \midrule
        & \multicolumn{1}{c}{SDA} & \multicolumn{1}{c}{Baseline} & \multicolumn{1}{c}{SDA} & \multicolumn{1}{c}{Baseline} & \multicolumn{1}{c}{SDA} & \multicolumn{1}{c}{Baseline} & \multicolumn{1}{c}{SDA} & \multicolumn{1}{c}{Baseline} \\
$IS_A$  & \textbf{3.051}                   & 2.802                         & \textbf{3.214}                 & 1.763                         & \textbf{3.799}                   & 2.681                         & 3.118                   & \textbf{3.256}                         \\
$IS_B$  & \textbf{2.928}                   & 2.913                         & \textbf{3.140}                  & 2.998                         & \textbf{4.254}                   & 2.962                         & \textbf{3.018}                   & 2.892                         \\
$FID_A$ & \textbf{53.5}                   & 142.8                         & \textbf{206.9}                   & 208.7                         & \textbf{168.3}                   & 171.2                         & \textbf{166.8}                   & 190.4                         \\
$FID_B$ & 103.1                   & \textbf{101.2}                        & 172.5                   & \textbf{171.0}                        & \textbf{174.3}                   & 174.7                         & \textbf{188.2}                   & 243.2                         \\ \bottomrule
\end{tabular}
\end{table*}

\subsection{Cycle-consistency loss}
In order to preserve translation consistency, we apply two independent mapping network $G:X\rightarrow Y$ and $H:Y\rightarrow X$. A full cycle in an iteration step can be described as follows:

Firstly, the generator $G$ transforms an input image $x$ from domain $X$ to domain $Y$. The generated result $G(x)$, together with the sampled image y from domain $Y$ are used to calculate the adversarial loss. Then $G(x)$ is processed by another generator $H$, which is responsible for translation from $Y$ to $X$. The output will be $H(G(x))$, which can be used to calculate cycle-consistency loss with input $x$. The distance ($L_1$, $L_2$, cross-entropy) between $H(G(x))$ and $x$ denotes the information loss after recurrently passing $x$ into 2 generators. The below equation formulate the cycle-consistency loss using Wasserstein loss \cite{arjovsky2017wasserstein}:
\begin{equation}
\begin{aligned}
\mathcal{L}_{\text {cycle }}(G, F) &=\mathbb{E}_{x \sim p_{\text {data }}(x)}\left[\|H(G(x))-x\|_{1}\right] \\
&+\mathbb{E}_{y \sim p_{\text {data }}(y)}\left[\|G(H(y))-y\|_{1}\right]
\end{aligned}
\end{equation}

The identity loss is used to value the consistency when the network is inputted with the image from the output domain, for example, $G(y)$. Intuitively this output should remain unchanged if the network is well trained. So the distance between$ G(y)$ and $y$, as well as $H(x)$ and $x$ is measured and added into the total loss. The identity loss is formulated as follows:
\begin{equation}
\begin{aligned}
\mathcal{L}_{\text {id}}(G, F) &=\mathbb{E}_{x \sim p_{\text {data }}(x)}\left[\|H(x)-x\|_{1}\right] \\
&+\mathbb{E}_{y \sim p_{\text {data }}(y)}\left[\|G(y)-y\|_{1}\right]
\end{aligned}
\end{equation}

\begin{figure*}[t]
    \centering
    \includegraphics[width=0.6\textwidth]{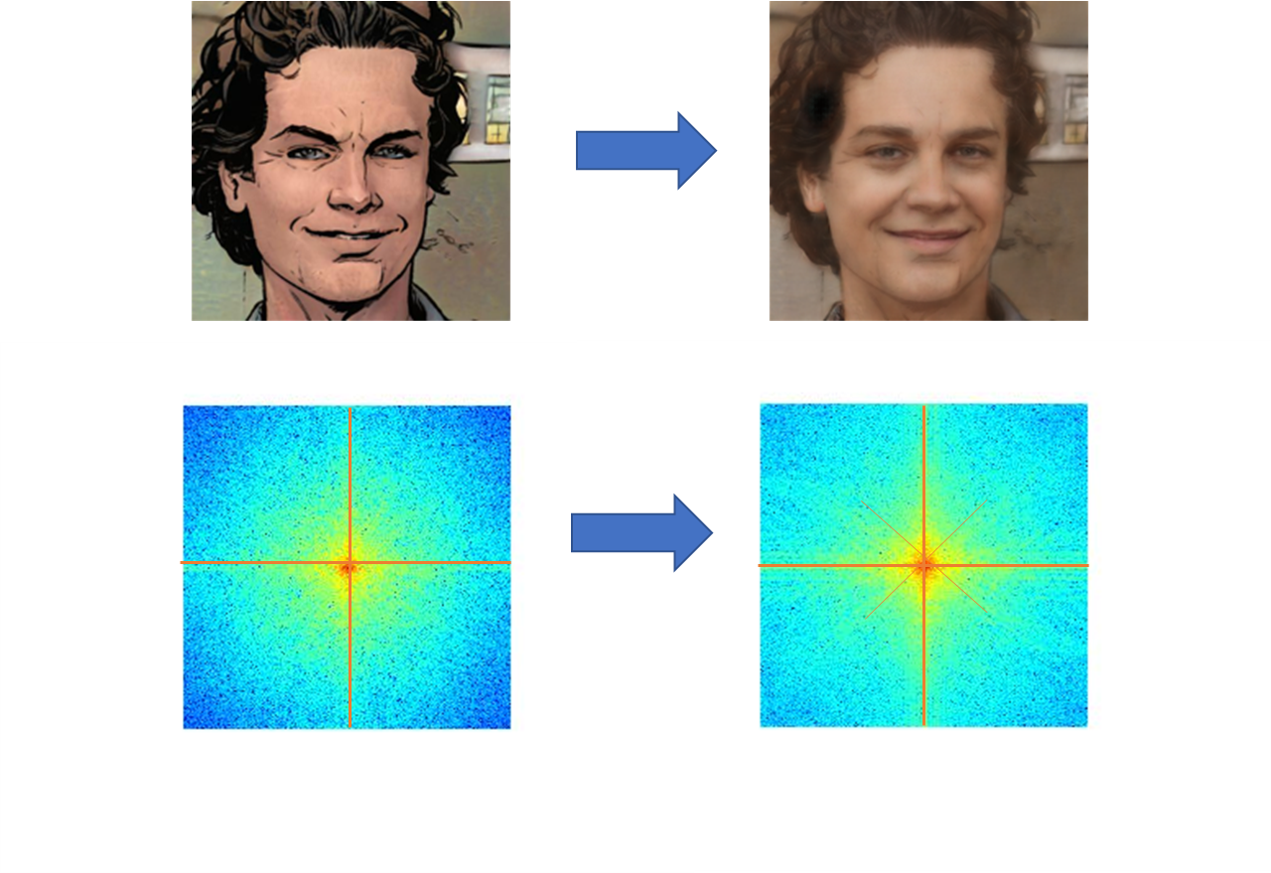}
    \caption{Spectral filling with SDA-GAN }
    \label{spe}
\end{figure*}

\section{Experimental analysis}
In order to illustrate the functionalllity of spectral attention, we choose the AttentionGAN from \cite{tang2019attentiongan} as the baseline network. We trained the proposed SDA-GAN in Fig.\ref{sda} (b) and the baseline in a variety of datasets including natural objects, comic styled images and Monet paintings. The networks share the same set of hyperparameters and are both trained in a single v100 GPU for 200 epoches. Part of the testing results are shown in Fig.\ref{res}.

As we can see, Fig.\ref{res}(a) compares the translation results in horse2zebra and apple2orange dataset. The images generated by the baseline have extreme points, while the images produced by our approach are more normal and smooth. This difference is mainly because of the use of image fuser. In the baseline model, the final output is a simple summation of branches, with attention as the weight of each content image. Instead, our model applies an multi-layered convolutional network to perform the final adjustment of the output. Since the kernels process all-channel feature maps in one convolution, attention and content layers from different domain can then have interaction. Possible extreme value that is wrongly produced by certain layer can be then eliminated by the more confident information from other layers. The interaction between attention and contents, we argue, is a more efficient way to utilize such feature.  Fig.\ref{res}(b) is the style-transform results in face2comic and monet2photo datasets. The baseline and SDA-GAN are both cycled-structure, so we performed the forward translation A-B and backward translation B-A in the all 4 datasets. The models are evaluated by Inception Score \cite{salimans2016improved} and FID metric \cite{heusel2017gans}. Detailed comparision is in Table.1:

The chart indicates that SDA-GAN has better IS compared with baseline network in the same training manner. Thus, we can conclude that our model can generate more balanced classes of image, and has less possibility of mode collapse. The FID metric, different from IS, is used to measure the distance between generated image and target domain. From the listed evaluations, the FID metrics of both model, in terms of real object translation (column 2 and 3) are similar. However in style transfer tasks, such as photo to Monet painting, the FID is largely reduced with our model which takes spectral attention. This is consistent with our hypothesis, that both object-scaled, high-level features and style-based, low-level  features can be generated with spectral engineering. SDA-GAN modifies the texture and color theme into the distribution of target domain. 

We also noticed that the FID of reconstructed human face images from comics has achieved a large improvement (from 142.8 to 53.5). Following the concept of spectral engineering, we analysed the 2D FFT of generated image, illustrated in Fig.\ref{spe}. Before translation, the comic picture only has two main frequency components, along the x-axis and y-axis. This indicates the dominance of horizontal and vertical in the source image. The translation introduces more dispersed spectral distribution, in the perspective of signal analysis, more high-frequency parts on the output picture. This phenomenon leads to the more 'realistic' profile of reconstructed face image.

\section{Conclusion}

This work introduced spectral engineering into image translation. Starting from comparing the spectra of real face picture and comic images, we formulated the high-level and low-level translation,by separating phase and amplitude components.

 Since Fast Fourier Transform (FFT) is based on the linear combination of input signals, it’s reasonable to assume convolutional network can function as a space-frequency domain converter naturally. Our approach pilot and accelerated the convergence to this potential function for spectral-related branches. Besides, little computational cost is imported in this approach, as the derivative of Fourier transform can be represented by inverse Fourier transform, which is, in the same level of complexity. 

Our proposed model is trained in limited time and with limited resources. For future study, we can increase the number of content layers of the decoder. This would introduce more freedom in putting attention on different semantic components. Another possible improvement can be made by adding spectral information to the discrimination process. In a word, we 'hard code' the spectral layer into the discriminator network. This modification will consider spectral distribution and spectral consistency, and might increase the output quality.

\newpage

\bibliographystyle{unsrtnat}
\bibliography{references}  
\end{document}